\documentclass[5p]{elsarticle}

\usepackage{lineno}
\usepackage[
	colorlinks=true,
	urlcolor=blue,
	linkcolor=green
]{hyperref}
\modulolinenumbers[5]

\journal{Neural Networks}

\usepackage{amsmath} 
\usepackage{amssymb}  
\usepackage[pdftex, dvipsnames]{xcolor}
\usepackage[utf8]{inputenc}
\usepackage{adjustbox}

\usepackage{etoolbox}
\makeatletter
\patchcmd{\ps@pprintTitle}
  {Preprint submitted to}
  {DOI: \url{https://doi.org/10.1016/j.neunet.2018.10.005} published in}
  {}{}
\makeatother

\newcommand{\figureFramework}{
\begin{figure*}[!htbp]
\centering 
 	\includegraphics[scale=1.4]{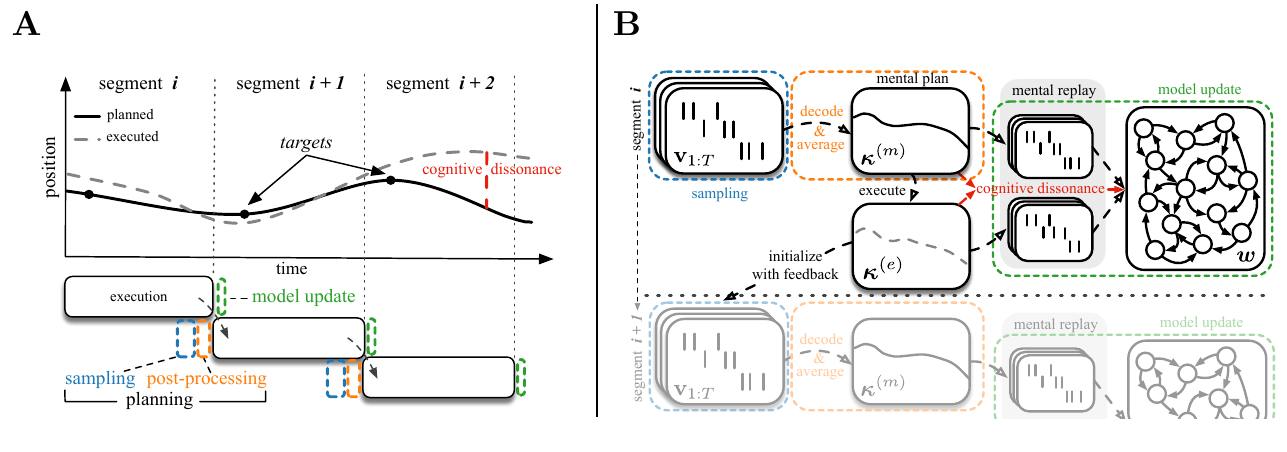}
 	\caption{
 	\textbf{Conceptual sketch of the framework.}
 	\textbf{A} shows the online planning and adaptation concept of using short segments. 
 	On the upper part the idea of cognitive dissonance is illustrated with a planned and executed trajectory.
 	The steps sampling and post-processing for a segment are timed such that they are performed during the end of the execution of the previous segment, whereas model adaptation is performed at the beginning of the segment execution. 
 	\textbf{B} shows the process with two segments in detail, including sampling of movements, decoding and averaging for creating the mental plan and the model update.
 	The executed segment provides feedback for planning the next segment and the matching mental and executed trajectory pairs are used for updating the model based on their cognitive dissonance.
} 
 	\label{fig:onlineFramework} 
\end{figure*}
}

\newcommand{\figureRobots}{
\begin{figure*}[!htbp]
	\centering 
	\includegraphics[scale=1.4]{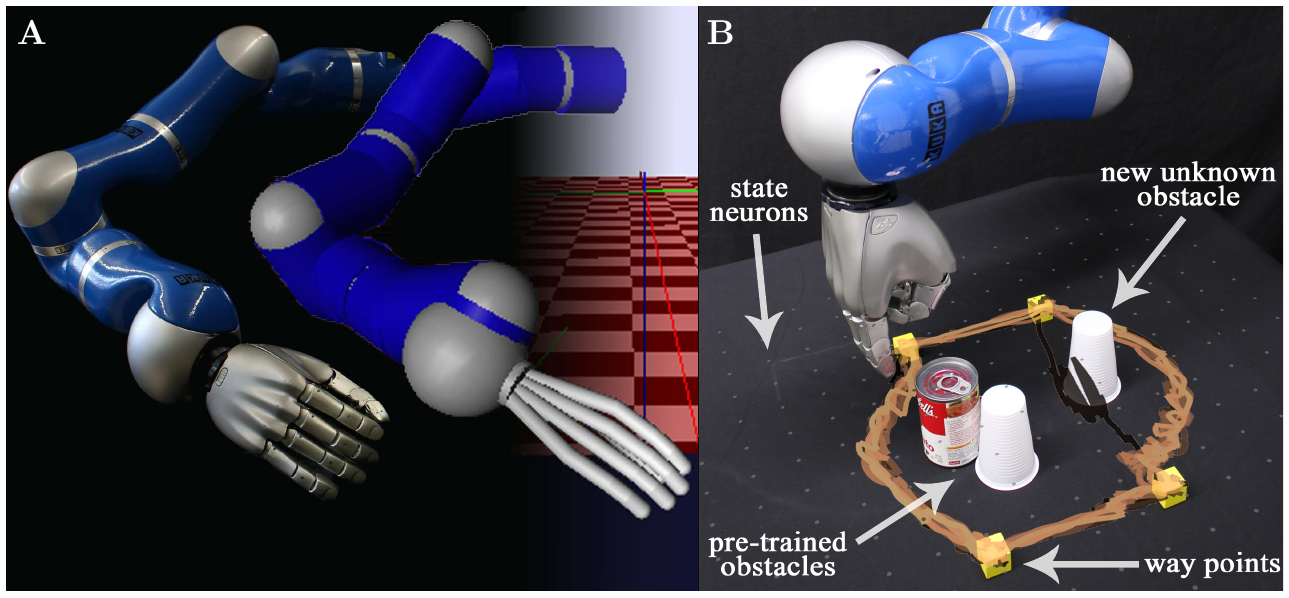}
	\caption{
	\textbf{Experimental setup.} 
	\textbf{A} shows the KUKA LWR arm (left) and its realistic dynamic simulation (right).
	\textbf{B} shows the setup for online learning on the real robot. 
	The model was initialized with one trial from the simulation of the robot (1st trial in Figure~\ref{fig:learningOverivewLocal}) and the new obstacle is learned additionally online on the real system. 
	The overlay shows the mental plan over one trial of about $5{:}30$ minutes. See Figure~\ref{fig:learningOverviewRobot} for more details.} 
	\label{fig:darias} 
\end{figure*}
}

\newcommand{\figureGlobal}{
\begin{figure*}[!tbp] 
	\centering 
		\includegraphics[scale=1.0]{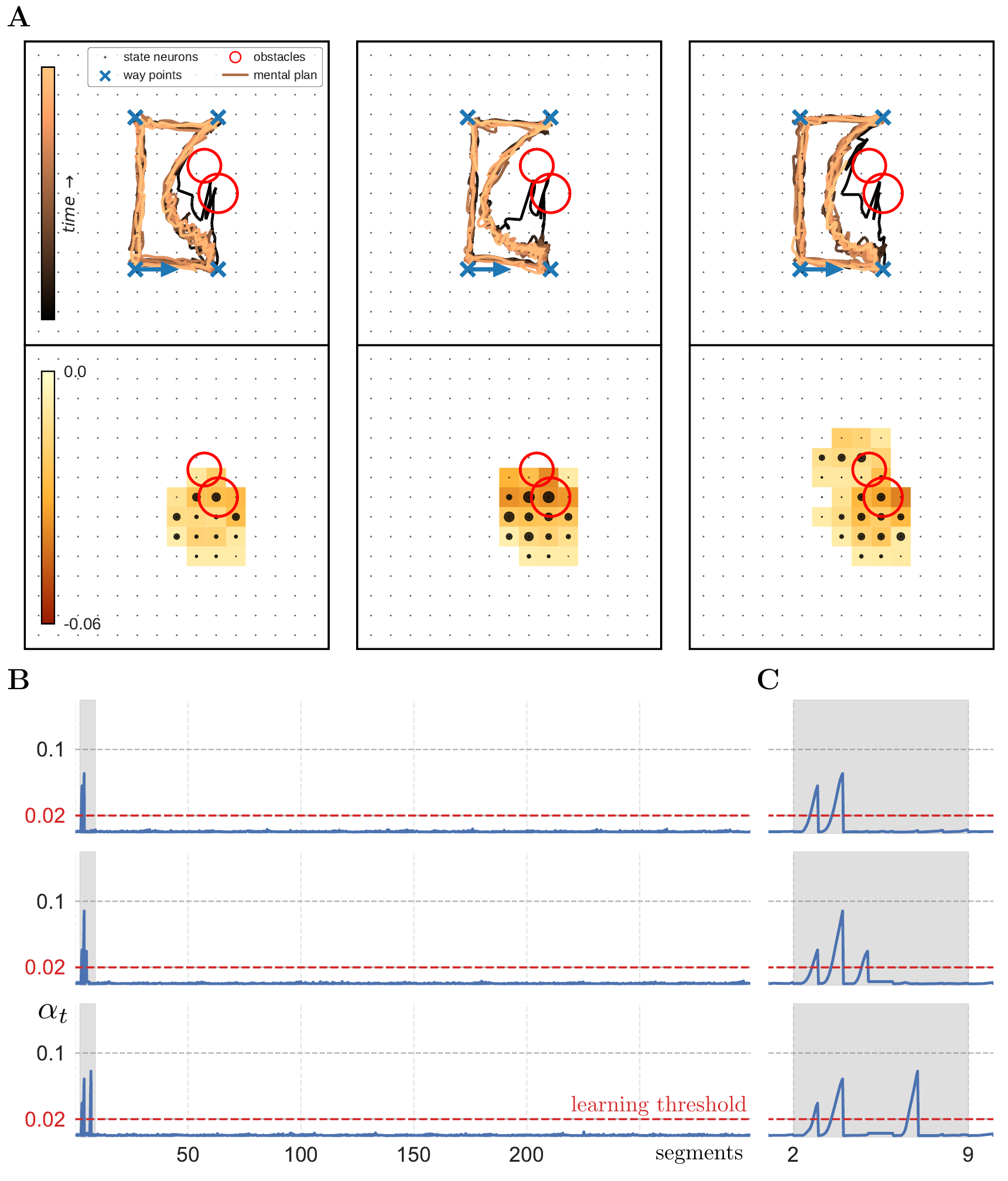}
			\caption{\textbf{Adaptation results for three trials with the global learning signal.}
			Each column in \textbf{A} shows one trial of the online adaptation with the \textit{global} learning signal, where the upper row shows the mental plan over time and the lower row depicts the adapted model.
			This change in the model is depicted with the heatmap showing the \textit{average change} of \textit{synaptic input} each neuron receives.
			Similarly the \textit{average change} of \textit{synaptic output} each neuron sends is shown with the scaled neuron sizes.
			\textbf{B} and \textbf{C} show the global learning signals $\alpha_t$ for the three trials over the planned segments.
			} 
			\label{fig:learningOverivewGlobal} 
\end{figure*}
}

\newcommand{\figureLocal}{
\begin{figure*}[!tbp] 
	\centering 
		\includegraphics[scale=1.0]{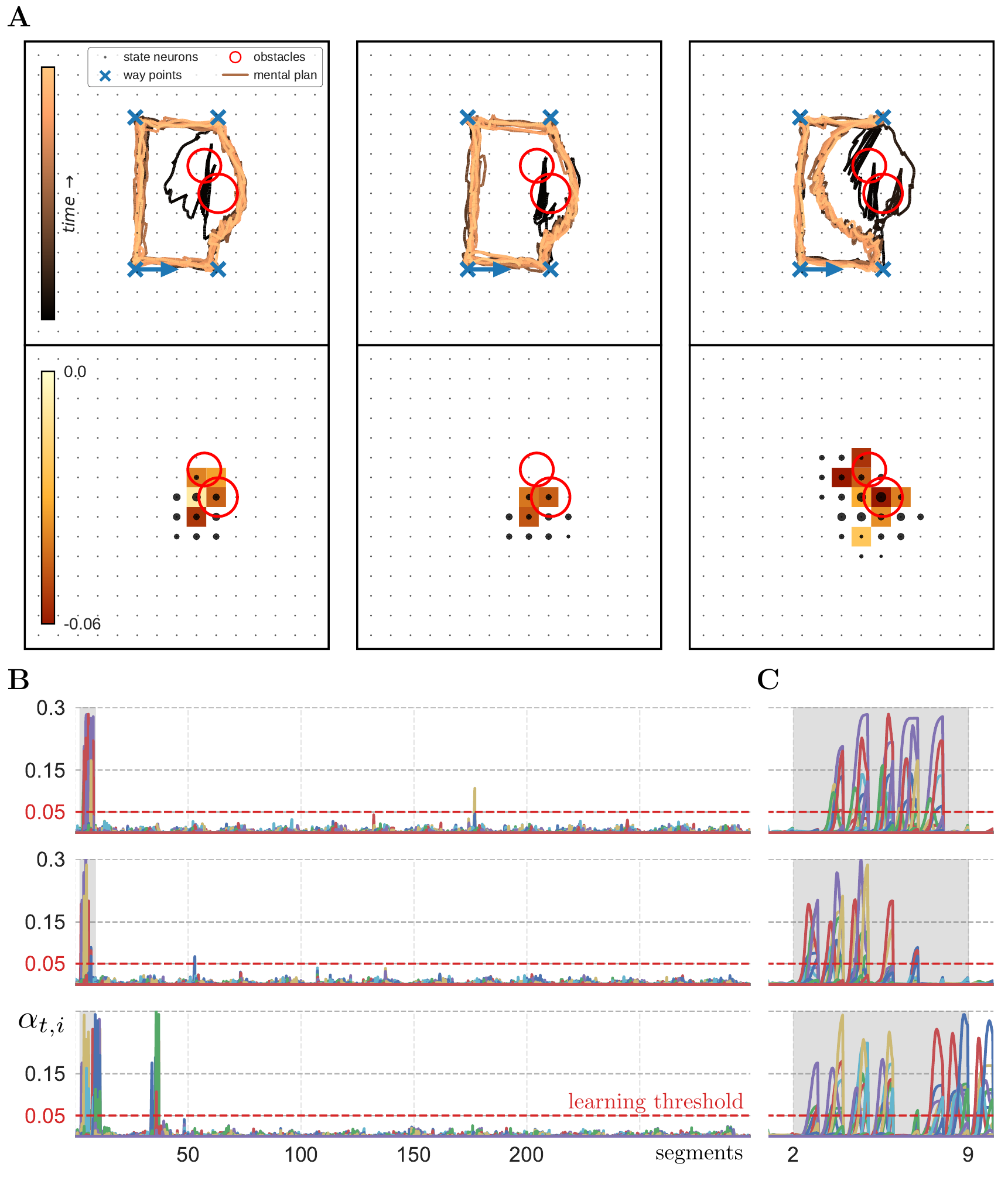}
			\caption{\textbf{Adaptation results for three trials with the local learning signals.} 
			Each column in \textbf{A} shows one trial of the online adaptation with the \textit{local} learning signals, where the upper row shows the mental plan over time and the lower row depicts the changed model.
			This change in the model is depicted with the heatmap showing the \textit{average change} of \textit{synaptic input} each neuron receives.
			Similarly the \textit{average change} of \textit{synaptic output} each neuron sends is shown with the scaled neuron sizes.
			\textbf{B} and \textbf{C} show the local learning signals $\alpha_{t,i}$ for the three trials over the planned segments.
			Each color indicates the learning signal for one neuron.
			}
			\label{fig:learningOverivewLocal} 
\end{figure*}
}

\newcommand{\figureTransfer}{
\begin{figure*}[!tbp] 
	\centering 
		\includegraphics[scale=1.0]{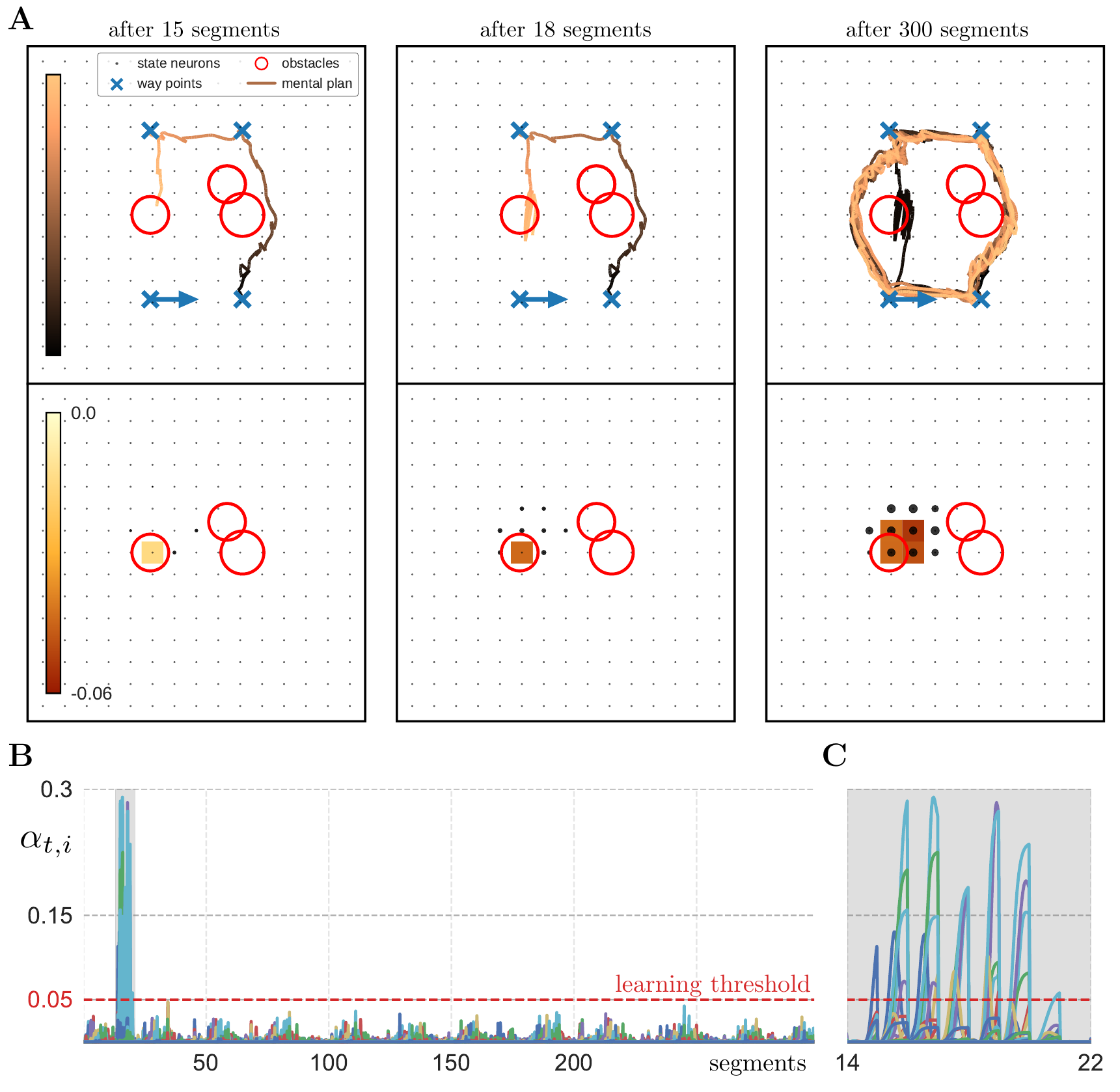}
			\caption{
			\textbf{Adaptation results on the real robot.}
			Online adaptation with the real KUKA LWR arm using the local learning signals initialized with simulation results, i.e., the right obstacles are already learned.
			The left obstacle is added to the real environment (see Figure~\ref{fig:darias}B).
			Each column in \textbf{A} shows the mental plan and the model for the indicated time.
			The change in the model is depicted with the heatmap showing the \textit{average change} of \textit{synaptic input} each neuron receives compared to the \textit{pre-trained model}.
			Similarly the \textit{average change} of \textit{synaptic output} each neuron sends is shown with the scaled neuron sizes.
			The mental plan demonstrates the rapid adaptation, as only a few interactions of the robot are necessary to adapt to the new environment.
			This efficiency is further highlighted in \textbf{B} and \textbf{C}, where the local learning signals $\alpha_{t,i}$ are shown over the execution time.
			Each color indicates the learning signal for one neuron.
			} 
			\label{fig:learningOverviewRobot} 
			\vspace{-10pt}
\end{figure*}
}

\newcommand{\figureCompare}{
\begin{figure*}[!tbp] 
	\centering 
		\includegraphics[scale=0.9]{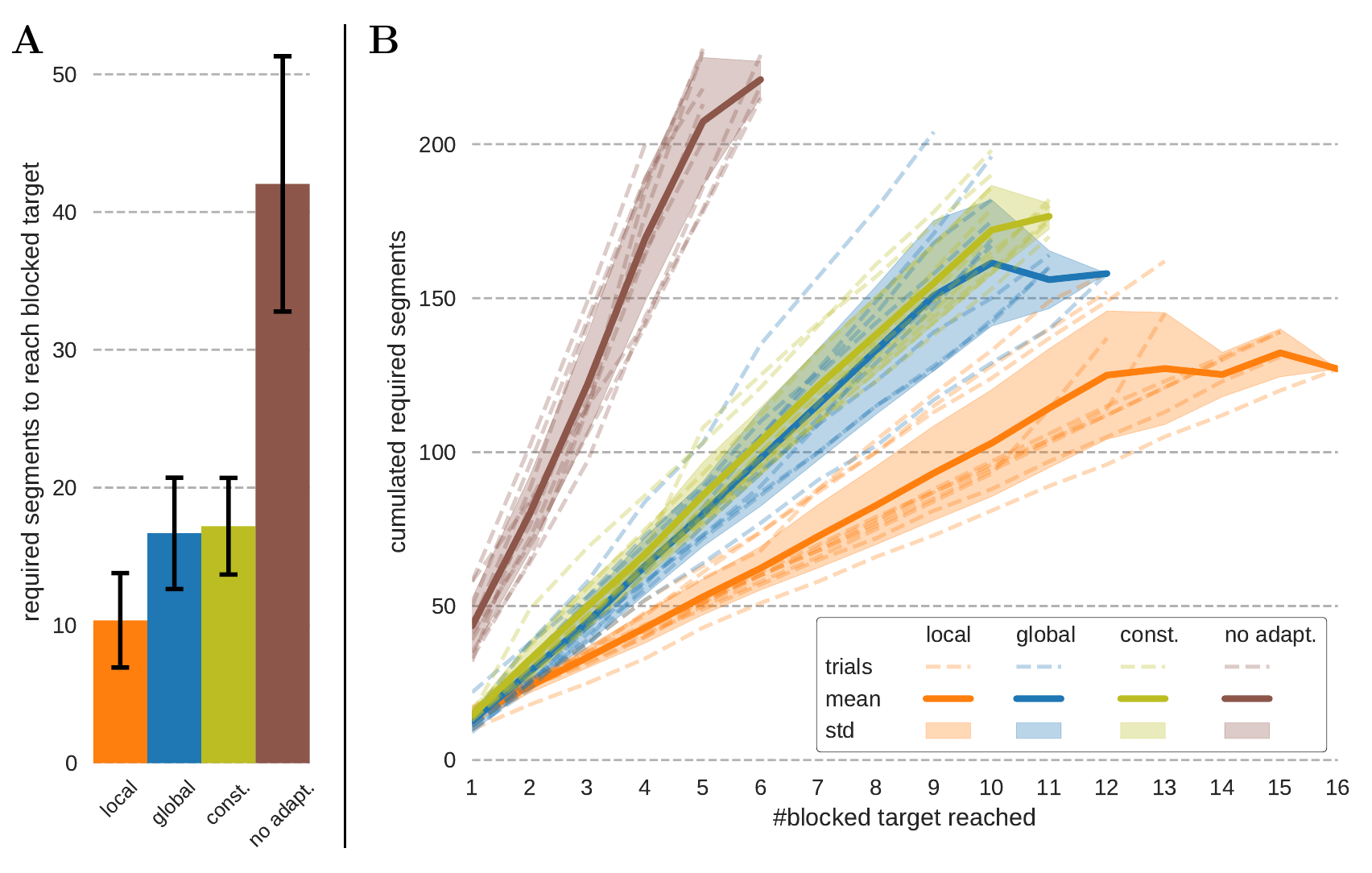}
			\caption{\textbf{Efficiency of the learned solutions.}
			\textbf{A} shows the mean and standard deviation of the number of required segments to reach the blocked target over all $10$ trials for each setting.
			\textbf{B} shows the cumulated required segments for reaching the blocked target for each trial and setting together with the mean and standard deviation of the trials of one setting.
			Note that due to the limit of planning $300$ segments in each trial, the number of times the blocked target is reached differs across trials. The constant learning rate ($\alpha = 0.001$) still uses the global adaptation signal for triggering learning.} 
			\label{fig:compare} 
\end{figure*}
}

\newcommand{\figureCompareHist}{
\begin{figure*}[!tbp] 
	\centering 
		\includegraphics[scale=0.9]{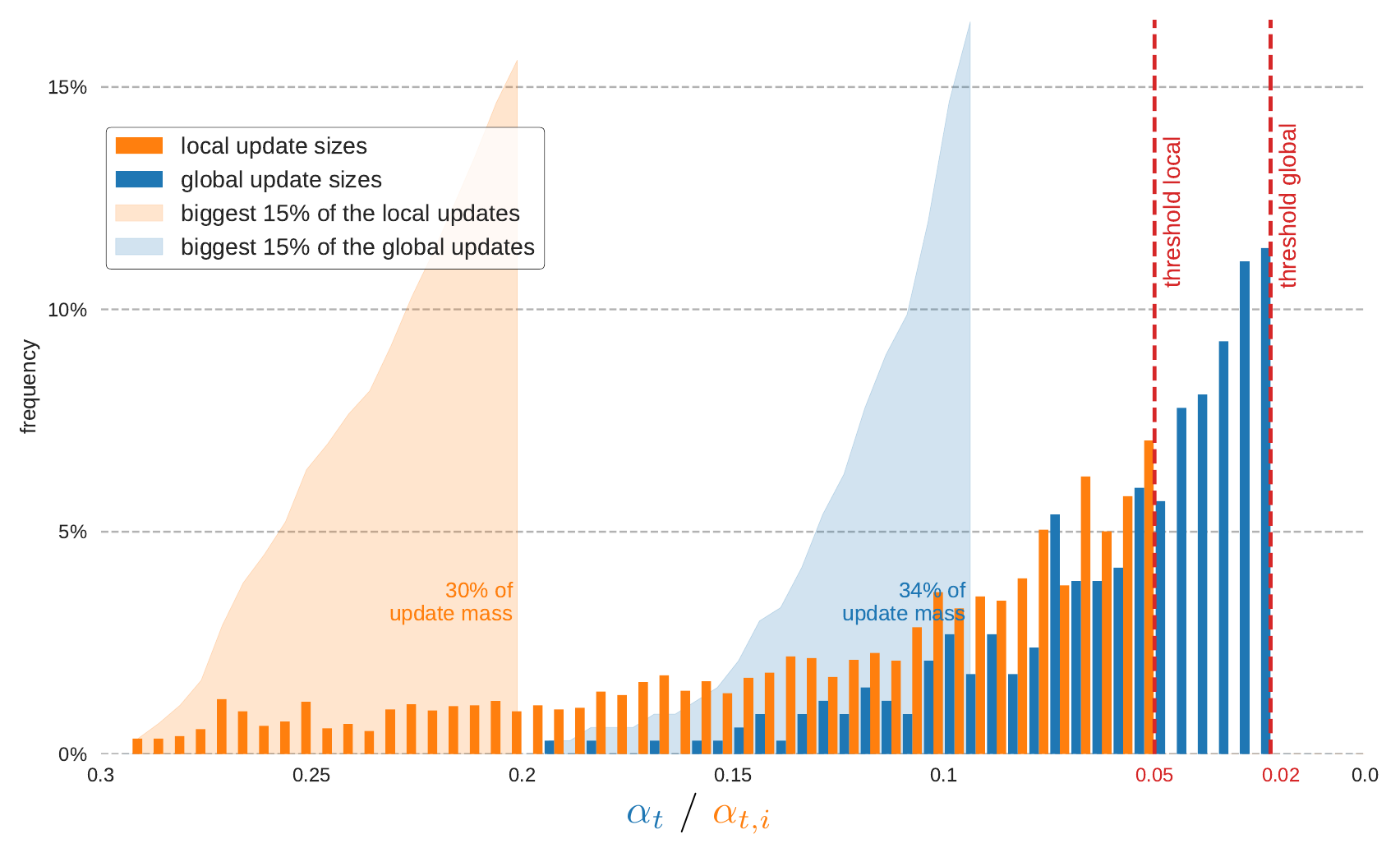}
			\caption{\textbf{Comparison of the learning signals.}
			The magnitude of the generated learning signals over all $10$ trials for each of the global and local mechanisms are shown with their respective frequencies.
			Update mass refers to the sum of all generated learning signals weighted by their frequencies.
			} 
			\label{fig:compareHist} 
\end{figure*}
}

\newcommand{\figureConst}{
\begin{figure*}[!tbp] 
	\centering 
		\includegraphics[scale=0.85]{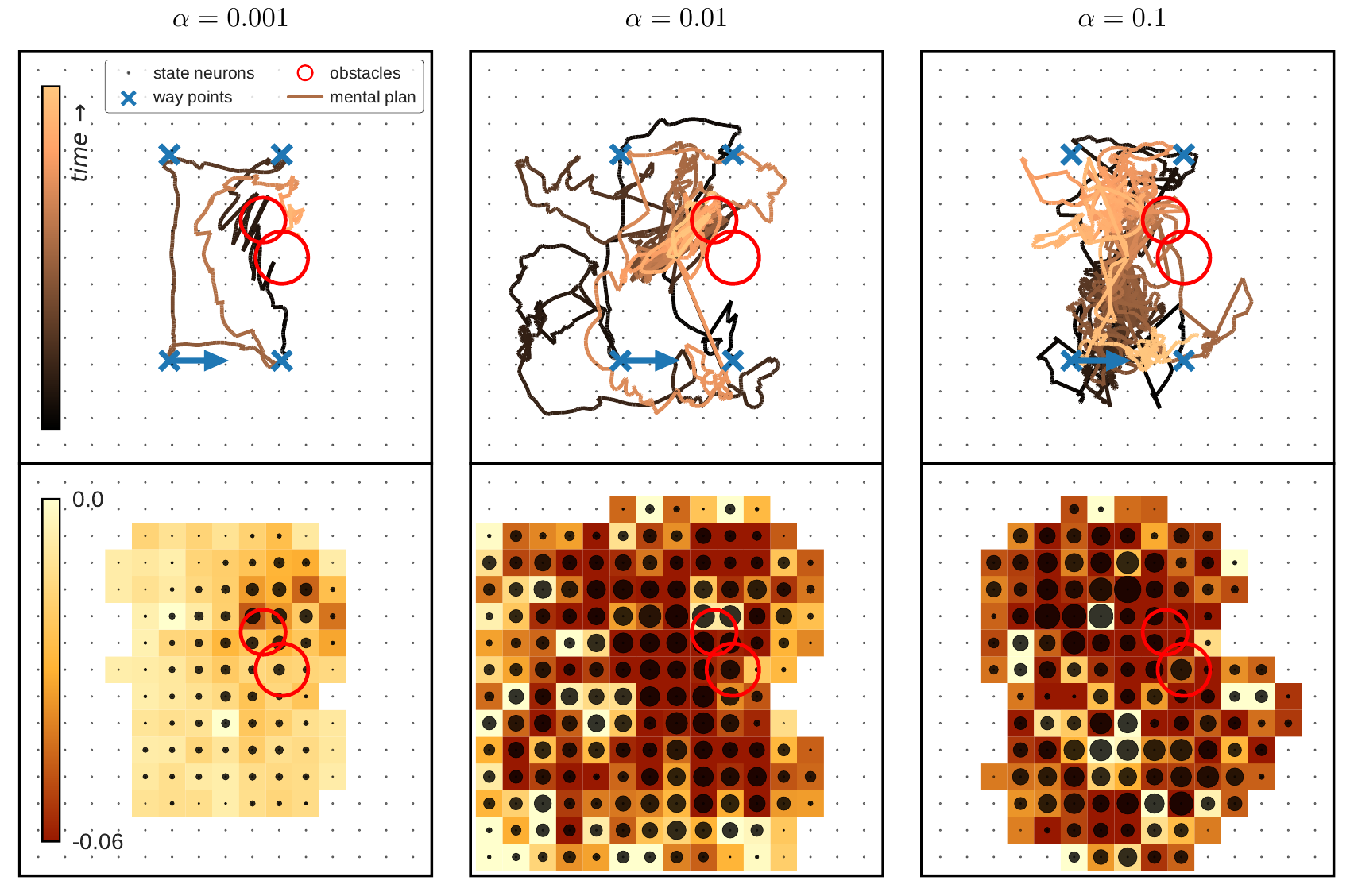}
			\caption{\textbf{Adaptation results with constant learning rates.}
			Each column shows one trial of the online adaptation with different constant learning rates $\alpha$, where the upper row shows the mental plan over time and the lower row depicts the adapted model.
			This change in the model is depicted with the heatmap showing the \textit{average change} of \textit{synaptic input} each neuron receives.
			Similarly the \textit{average change} of \textit{synaptic output} each neuron sends is shown with the scaled neuron sizes.
			With constant learning rates and no adaptive signal for triggering learning, the model is updated constantly, by what the state transition model gets destroyed and no correct movement can be sampled anymore.	
			} 
			\label{fig:learningOverivewFixed} 
\end{figure*}
}

\begin{document}

\begin{frontmatter}
\title{Intrinsic Motivation and Mental Replay enable\\Efficient Online Adaptation in Stochastic Recurrent Networks}

\author[label1]{Daniel Tanneberg\corref{cor1}}
\ead{daniel@robot-learning.de}
\author[label1,label2]{Jan Peters}
\ead{mail@jan-peters.net}
\author[label3,label1]{Elmar Rueckert}
\ead{rueckert@rob.uni-luebeck.de} 

\cortext[cor1]{Corresponding author}

\address[label1]{Intelligent Autonomous Systems, Technische Universit\"at Darmstadt,\\ Hochschulstr. 10, 64289 Darmstadt, Germany}
\address[label2]{Robot Learning Group, Max-Planck Institute for Intelligent Systems,\\ Max-Planck-Ring 4, 72076 T\"ubingen, Germany}
\address[label3]{Institute for Robotics and Cognitive Systems, Universität zu L\"ubeck,\\Ratzeburger Allee 160, 23538 L\"ubeck, Germany}

\nonumnote{\textcopyright 2018. Licensed under the Creative Commons CC-BY-NC-ND 4.0 license http://creativecommons.org/licenses/by-nc-nd/4.0/}

\begin{abstract}
Autonomous robots need to interact with unknown, unstructured and changing environments, constantly facing novel challenges.
Therefore, continuous online adaptation for lifelong-learning and the need of sample-efficient mechanisms to adapt to changes in the environment, the constraints, the tasks, or the robot itself are crucial.
In this work, we propose a novel framework for probabilistic online motion planning with online adaptation based on a bio-inspired stochastic recurrent neural network. 
By using learning signals which mimic the intrinsic motivation signal \textit{cognitive dissonance} in addition with a mental replay strategy to intensify experiences, the stochastic recurrent network can learn from few physical interactions and adapts to novel environments in seconds. 
We evaluate our online planning and adaptation framework on an anthropomorphic KUKA LWR arm. 
The rapid online adaptation is shown by learning unknown workspace constraints sample-efficiently from few physical interactions while following given way points. 
\end{abstract}


\end{frontmatter}

\figureFramework
\section{Introduction}
One of the major challenges in robotics is the concept of developmental robots~\cite{lungarella2003developmental, schmidhuber2006developmental,asada2009cognitive}, i.e., robots that develop and adapt autonomously through lifelong-learning~\cite{thrun1995lifelong,ring1997child,ruvolo2013ella}.
Although a lot of research has been done for learning tasks autonomously in recent years, experts with domain knowledge are still required in many setups to define and guide the learning problem, e.g., for reward shaping, for providing demonstrations or for defining the tasks that should be learned.
In a fully autonomous self-adaptive robot however, these procedures should be carried out by the robot itself.
In other words, the robot and especially its development should not be limited by the learning task specified by the expert, but should rather be able to develop \textit{on its own}.
Thus, the robot should be equipped with mechanisms enabling autonomous development to \textit{understand} and decide when, what, and how to learn~\cite{weng2001autonomous,weng2004developmental}.

Furthermore, as almost all robotic tasks involve movements and therefore movement planning, this developing process should be continuous.
In particular, planning a movement, executing it, and learning from the results should be integrated in a continuous online framework.
This idea is investigated in iterative learning control approaches~\cite{tayebi2004adaptive,bristow2006survey}, which can be seen as a simple adaptation mechanism that learns to track given repetitive reference trajectories.
More complex adaptation strategies are investigated in model-predictive control approaches~\cite{gu2002neural,krause2012stabilization,camacho2013model,ibanez2014emergence} that simultaneously plan, execute and re-plan motor commands.
However, the used models are fixed and cannot adapt straightforwardly to new challenges. 

Online learning with real robots was investigated in~\cite{jamone2012autonomous}, where multiple models were learned online for reaching tasks.
Online learning of push recovery actions during walking in a humanoid robot was shown in~\cite{yi2011online}, and in~\cite{hersch2008online} a mechanism for online learning of the body structure of a humanoid robot was discussed.
Recurrent neural networks were used to learn body mappings in a humanoid robot~\cite{reinhart2011neural}, and for efficient online learning of feedback controllers~\cite{waegeman2012feedback}.
However, in all these online learning settings, the learning problem was designed and specified a priori by a human expert, providing extrinsic reward.

From autonomous mental development in humans however, it is known that intrinsic motivation is a strong factor for learning~\cite{ryan2000intrinsic,ryan2000self}.
Furthermore, intrinsically motivated behavior is crucial for gaining the competence, i.e., a set of reusable skills, to enable autonomy~\cite{white1959motivation}.
Therefore, the abstract concept of intrinsically motivated learning has inspired many studies in artificial and robotic systems, e.g. \cite{barto2004Intrinsically,baldassarre2011What,baldassarre2013Intr}, which investigate intrinsically motivated learning in the reinforcement learning framework~\cite{sutton1998reinforcement}.
Typically, such systems learn the consequences of actions and choose the action that maximizes a novelty or prediction related reward signal~\cite{stout2005intrinsically,nehmzow2013novelty,santucci2013intrinsic}.

Intrinsic motivation is used for self-generating reward signals that are able to guide the learning process without an extrinsic reward that has to be manually defined and provided by an expert.
For the concept of lifelong-learning, intrinsic motivation signals are typically used for incremental learning within hierarchical reinforcement learning~\cite{barto2003recent} and the options framework~\cite{sutton1999between}.
Starting with a developmental phase, the robots learn incrementally more complex tasks utilizing the previously and autonomously learned skills.
Furthermore, the majority of related work on intrinsically motivated learning focuses on concepts and simulations, and only few applications to real robotic systems exist, for example~\cite{oudeyer07Intr,hart2011learning}.

\figureRobots
\paragraph*{Contribution} 
The contribution of this work is a neural-based framework for robot control that enables efficient online adaptation during motion planning tasks.
A novel intrinsically motivated \textit{local} learning signal is derived and combined with an experience replay strategy to enable efficient online adaptation.
We implement the adaptation approach into a biologically inspired stochastic recurrent neural network for motion planning~\cite{rueckert16Recu,tanneberg16Deep}.
This work builds on recent prior studies where a \textit{global} learning signal was investigated~\cite{tanneberg2017online,tanneberg2017efficient}.
These global and local learning signals enable efficient task-independent online adaptation without an explicit specified objective or learning task.
In robotic experiments we evaluate and compare these global and local learning signals and discuss their properties.
This study shows that our framework is suitable for model based robot control tasks where adaptation of the state transition model to dynamically changing environmental conditions is necessary. 

The task-independent online adaptation is done by updating the recurrent synaptic weights encoding the state transition model. 
The proposed learning principle, therefore, can be applied to model-based (control) approaches with internal (transition) models, like, for example, (stochastic) optimal control~\cite{kappen2012optimal,botvinick2012planning,ruckert2012learned} and model-predictive control~\cite{gu2002neural,krause2012stabilization,camacho2013model,ibanez2014emergence}.  
Furthermore, the method is embedded into a novel framework for continuous online motion planning and learning that combines the scheduling concept of model-predictive control with the adaptation idea of iterative learning control.

The online model adaptation mechanism uses a supervised learning approach and is modulated by intrinsic motivation signals that are inspired by cognitive dissonance~\cite{festinger1962cognitive,kagan1972motives}.
We use a knowledge-based model of intrinsic motivation~\cite{oudeyer2007what} that describes the divergence of the expectation to the observation.
This intrinsic motivation signal tells the agent where its model is \textit{incorrect} and guides the adaptation of the model with this mismatch.
In our experiments, this dissonance signal relates to a tracking error, however, the proposed method is more general and can be used with various modalities like vision or touch.
We derive two different mechanisms to compute the dissonance, a \textit{global} learning signal that captures the \textit{distance} between mental and executed trajectory, and a \textit{local} learning signal that takes the neurons \textit{responsibilities} for encoding these trajectories into account. 
These learning signals trigger the online adaptation when \textit{necessary} and guide the strength of the update.

Additionally, to intensify the effect of the experience, we use a mental replay mechanism, what has been proposed to be a fundamental concept in human learning~\cite{foster2006reverse}.
This mental replay is implemented by exploiting the stochastic nature of the spiking neural network model and its spike encodings of trajectories to generate multiple sample encodings for every experienced situation.

We will show that the stochastic recurrent network can adapt efficiently to novel environments without specifying a learning task within seconds from few interactions by using the proposed intrinsic motivation signals and a mental replay strategy on a simulated and real robotic system (shown in Figure~\ref{fig:darias}).

\subsection{Related Work on Intrinsically Motivated Learning}
In this subsection we discuss the related work for intrinsically motivated learning from practical and theoretical perspectives.

Early work on intrinsically motivated learning not using the typically reinforcement learning framework used the prediction error of sensory inputs for self-localization tasks~\cite{herrmann2000learning}.
In an online setup, the system explored novel and interesting stimuli to learn a representation of the environment.
By using this intrinsic motivation signal, the system developed structures for perception, representation and actions in a neural network model.
Actions were chosen such that the expected increase of knowledge was maximized.
The approach was evaluated in a gridworld domain and on a simple mobile robot platform.

Intrinsic motivation signals prediction, familiarity (in terms of frequency of state transitions) and stability (in terms of sensor signals to its average) were investigated in~\cite{kaplan2003motivational} in task-independent online visual exploration problems in simulation and on a simple robot.

By using the hierarchical reinforcement learning framework and utilizing the intrinsic motivation signal novelty, autonomous learning of a hierarchical skill collection in a playroom simulation was shown in~\cite{barto2004Intrinsically}. 
The novelty signal directed the agent to novel situations when it got \textit{bored}.
As already learned skills can be used as actions in new policies, the approach implements an incremental learning setup.

A similar approach was investigated in~\cite{hart2011learning}, were a framework for lifelong-learning was proposed.
This framework learns hierarchical polices and has similarities to the options framework.
By implementing a motivation signal based on affordance discovery\footnote{Affordance refers to the possibility of applying actions to objects or the environment.}, a repertoire of movement primitives for object detection and manipulation was learned on a platform with two robotic arms.
The authors also showed that these primitives can be sequenced and generalized to enable more complex and robust behavior.

Another approach for lifelong-learning based on hierarchical reinforcement learning and the options framework is shown in~\cite{metzen2013incremental}.
The authors learn incrementally a collection of reusable skills in simulations, by implementing the motivation signals novelty for learning new skills and prediction error for updating existing skills.

A different approach based on competence improvement with hierarchical reinforcement learning is discussed in~\cite{stout2010competence}.
The agent is given a set of skills, or options as in the options framework, and needs to choose which skill to improve.
The used motivation signal competence is implemented as the expected return of a skill to achieve a certain goal.
Rewards are generated based on this competence progress and the approach is evaluated in a gridworld domain.

In~\cite{oudeyer07Intr}, the \textit{intelligent adaptive curiosity} system is introduced and used to lead a robot to maximize its learning progress, i.e., guiding the robot to situations, that are neither too predictable nor too unpredictable.
The reinforcement learning problem is simplified to only trying to maximize the expected reward at the next timestep and a positive reward is generated when the error of an internal predictive model decreases.
Thus, the agent focuses on exploring situations whose complexity matches its current abilities.
The mechanism is used on a robot that learns to manipulate objects.
The idea is to equip agents with mechanisms \textit{computing} the degree of novelty, surprise, complexity or challenge from the robots point of view and use these signals for guiding the learning.

In~\cite{santucci2013intrinsic} different prediction based signals are investigated within a reinforcement learning framework on a simulated robot arm learning reaching movements.
The framework uses multiple expert neural networks, one for each task, and a selection mechanism that determines which expert to train.
The motivation signals are implemented with learned predictors with varying input that learn to predict the achievement of the selected task.
Predicting the achievement of the task once in the beginning of a trial produced the best results.

Recently, open-ended learning systems based on intrinsic motivation increasingly give importance to explicit goals -- known from the idea of goal babbling for learning inverse kinematics~\cite{Rolf2010Goalbabbling} -- for autonomous learning of skills to manipulate the robots environment~\cite{Santucci2016GRAIL}.

Beside the aforementioned more practical research, also work on theoretical aspects of intrinsic motivated learning exists.
For example, a coherent theory and fundamental investigation of using intrinsic motivation in machine learning over two decades is discussed in~\cite{schmidhuber2010formal}.
The authors state that the improvement of prediction errors can be used as an intrinsic reinforcement for efficient learning.

Another comprehensive overview of intrinsically motivated learning systems is given in~\cite{baldassarre2013Intr}.
The authors introduce three classes for clustering intrinsic motivation mechanisms.
In particular, they divide these mechanisms into prediction based, novelty based and competence based approaches, and discuss their features in detail.
Furthermore, that prediction based and novelty based intrinsic motivations are subject to distinct mechanisms was shown in~\cite{barto2013Novelty}.

In~\cite{oudeyer2007what} a psychological view on intrinsic motivation is discussed and a formal typology of computational approaches for studying such learning systems is presented.

Typically intrinsic motivation signals have been used for incremental task learning, acquiring skill libraries, learning perceptual patterns and for object manipulation.
For the goal of fully autonomous robots however, the ability to focus and guide learning independently from tasks, specified rewards and human input is crucial.
The robot should be able to learn without \textit{knowing} what it is supposed to learn in the beginning.
Furthermore, the robot should detect on its own if it needs to learn something new or adapt an existing ability if its internal model differs from the perceived reality.
To achieve this, we equip the robot with a mechanism for task-independent online adaptation utilizing intrinsic motivation signals inspired by cognitive dissonance.
For rapid online adaptation within seconds, we additionally employ a mental replay strategy to intensify experienced situations.
Adaptation is done by updating the synaptic weights in the recurrent layer of the network that encodes the state transition model, and this learning is guided by the cognitive dissonance inspired signals.

\section{Materials and Methods}
In this section, we first summarize the challenge and goal we want to address with this paper.
Afterwards, we describe the functionality and principles of the underlying bio-inspired stochastic recurrent neural network model, that samples movement trajectories by simulating its inherent dynamics.
Next we introduce our novel framework, which enables this model to plan movements online and show how the model can adapt online utilizing intrinsic motivation signals within a supervised learning rule and a mental replay strategy.
 
\subsection{The Challenge of (Efficient) Online Adaptation in Stochastic Recurrent Networks}
The main goal of the paper is to show that efficient online adaptation of stochastic recurrent networks can be achieved by using intrinsic motivation signals and mental replay.
Efficiency is measured as the number of updates triggered, which is equal to the number of required samples, e.g., here the number of physical interactions of the robot with the environment.
Additionally, we will show that using adaptive learning signals and only trigger learning when necessary are crucial mechanisms for  updating such sensitive stochastic networks.

\subsection{Motion Planning with Stochastic Recurrent Neural Networks}
The proposed framework builds on the model recently presented in~\cite{rueckert16Recu}, where it was shown that stochastic spiking networks can solve motion planning tasks optimally.
Furthermore, in~\cite{tanneberg16Deep} an approach to scale these models to higher dimensional spaces by introducing a factorized population coding and that the model can be trained from demonstrations was shown.
 
Inspired by neuroscientific findings on the mental path planning of rodents~\cite{pfeiffer2013hippocampal}, the model mimics the behavior of hippocampal place cells.
It was shown that the neural activity of these cells is correlated not only with actual movements, but also with future mental plans.
This bio-inspired motion planner consists of stochastic spiking neurons forming a multi-layer recurrent neural network.
It was shown that spiking networks can encode arbitrary complex distributions~\cite{buesing2011neural} and learn temporal sequences~\cite{brea2011sequence,kappel2014stdp}.
We utilize these properties for motion planning and learning as well as to encode multi-modal trajectory distributions that can represent multiple solutions to planning problems.

The basis model consists of two different types of neuron populations: a layer of $K$ \textit{state} neurons and a layer of $N$ \textit{context} neurons.
The state neurons form a fully connected recurrent layer with synaptic weights $w_{i,k}$, while the context neurons provide feedforward input via synaptic weights $\theta_{j,k}$, with $j \in N$ and $k,i \in K$ with $N \ll K$.
There are no lateral connections between context neurons.
Each constraint or any task-related information is modeled by a population of context neurons. 
While the state neurons are uniformly spaced within the modeled state space, the \textit{task-dependent} context neurons are Gaussian distributed \textit{locally} around the corresponding location they encode, i.e., there are only context neurons around the specific constraint they encode. 

The state neurons can be seen as an abstract and simplified version of place cells and encode a cognitive map of the environment~\cite{stachenfeld2014design}. 
They are modeled by stochastic neurons which build up a membrane potential based on the weighted neural input. 
Context neurons have no afferent connections and spike with a fixed time-dependent probability. 
Operating in discrete time and using a fixed refractory period of $\tau$ timesteps that decays linearly, the neurons spike in each time step with a probability based on their membrane potential. 
All spikes from presynaptic neurons get weighted by the corresponding synaptic weight and are integrated to an overall postsynaptic potential (PSP). 
Assuming linear dendritic dynamics, the membrane potential of the state neurons is given by
\begin{linenomath}
\begin{align}
\label{eq:memPot}
	u_{t,k} = \sum_{i=1}^{K} w_{i,k}\tilde{v}_i(t) + \sum_{j=1}^{N} 
	\theta_{j,k}\tilde{y}_j(t) \  ,
\end{align} 
\end{linenomath}
where $\tilde{v}_i(t)$ and $\tilde{y}_j(t)$ denote the presynaptic input injected from neurons $i \in K$ and $j \in N$ at time $t$ respectively.
Depending on the used PSP kernel for integrating over time, this injected input can include spikes from multiple previous timesteps.
This definition implements a simple stochastic spike response model~\cite{gerstner2002spiking}.
Using this membrane potential, the probability to spike for the state neurons can be defined by $\rho_{t,k} = p(v_{t,k} = 1) = f(u_{t,k})$, where $f(\cdot)$ denotes the activation function, that is required to be differentiable.
The binary activity of the state neurons is denoted by $\mathbf{v}_t = (v_{t,1},..,v_{t,K})$, where $v_{t,k} = 1$ if neuron $k$ spikes at time t and $v_{t,k} = 0$ otherwise.
Analogously, $\mathbf{y}_t$ describes the activity of the context neurons. 
The synaptic weights $\boldsymbol{\theta}$ which connect context neurons to state neurons provide task related information.
By injecting this task related information, the context neurons modulate the random walk behavior of the state neurons towards goal directed movements.
This input from the context neurons can also be learned~\cite{rueckert16Recu} or can be used to, for example, include known dynamic constraints in the planning process~\cite{tanneberg16Deep}.

We compared setting the feedfoward context neuron input weights $\boldsymbol{\theta}$ as in~\cite{tanneberg16Deep} -- proportional to the euclidean distance -- to using Student's t-distributions and generalized error distributions, where the latter produced the best results and was used in the experiments.
At each context neuron position such a distribution is located and the weights to the state neurons are drawn from this distribution using the distance between the connected neurons as input.
For way points, these context neurons install a \textit{gradient} towards the associated position such that the random walk samples are biased towards the active locations.

For planning, the stochastic network encodes a distribution 
\begin{linenomath} 
\begin{align}
	q(\mathbf{v}_{1:T} | \boldsymbol{\theta}) = p(\mathbf{v}_0) \prod_{t=1}^{T} \mathcal{T}(\mathbf{v}_t | \mathbf{v}_{t-1}) \phi_t(\mathbf{v}_t | \boldsymbol{\theta}) \ \nonumber
\end{align}
\end{linenomath}
over state sequences ($\mathbf{v}_{1:T}$) of $T$ timesteps, where $\mathcal{T}(\mathbf{v}_t | \mathbf{v}_{t-1})$ denotes the transition model and $\phi_t(\mathbf{v}_t | \boldsymbol{\theta})$ the task related input provided by the context neurons.
Using the definition of the membrane potential from Equation~\eqref{eq:memPot}, the state transition model is given by
\begin{linenomath}
\begin{align}
\label{eq:transModel}
\mathcal{T}(v_{t,i} | \mathbf{v}_{t-1}) &= 
	f \left( \sum_{k=1}^{K} w_{k,i} \tilde{v}_k(t) v_{t,i} \right) \ ,
\end{align}
\end{linenomath}
where a PSP kernel that covers multiple time steps includes information provided by spikes from multiple previous time steps.
In particular, we use a rectangular PSP kernel of $\tau$ timesteps, given by
\begin{linenomath}
\begin{align}
	\tilde{v}_k(t) = \begin{cases}1 \text{ if } \exists l \in [t-\tau,t-1] : v_{l,k} = 1 
	\\0 \text{ otherwise}\end{cases} \ \nonumber,
\end{align}
\end{linenomath}
such that, if neuron $k$ has spiked within the last $\tau$ timesteps, the presynaptic input $\tilde{v}_k(t)$ is set to $1$.
Movement trajectories can be sampled by simulating the dynamics of the stochastic recurrent network~\cite{buesing2011neural} where multiple samples are used to generate smooth trajectories. 

\paragraph{Encoding continuous domains with binary neurons} 
All neurons have a preferred position in a specified coordinate system and encode binary random variables (spike $=1$/no spike $=0$). 
Thus, the solution sampled from the model for a planning problem is the spiketrain of the state neurons, i.e., a sequence of binary activity vectors. 
These binary neural activities encode the continuous system state $\mathbf{x}_t$, e.g., end-effector position or joint angle values,  using the decoding scheme 
\begin{linenomath}
\begin{align}
\mathbf{x}_t = \frac{1}{|\hat{\mathbf{v}}_t|} \sum_{k=1}^{K} \hat{v}_{t,k}\mathbf{p}_k \quad \text{with} \quad
 |\hat{\mathbf{v}}_t| = \sum_{k=1}^{K} \hat{v}_{t,k} \ , \nonumber  
\end{align}
\end{linenomath}
where $\mathbf{p}_k$ denotes the preferred position of neuron $k$ and $\hat{v}_{t,k}$ is the continuous activity of neuron $k$ at time $t$ calculated by filtering the binary activity $v_{t,k}$ with a Gaussian window filter.
Together with the dynamics of the network, that allows multiple state neurons being active at each timestep, this encoding enables the model to work in continuous domains.
To find a movement trajectory from position $\mathbf{a}$ to a target position $\mathbf{b}$, the model generates a sequence of states encoding a task fulfilling trajectory.

\subsection{Online Motion Planning Framework}
For efficient online adaptation, the model should be able to react during the execution of a planned trajectory.
Therefore, we consider a short time horizon instead of planning complete movement trajectories over a long time horizon.
This short time horizon sub-trajectory is called a \textit{segment}.
A trajectory $\boldsymbol{\kappa}$ from position $\mathbf{a}$ to position $\mathbf{b}$ can thus consist of multiple segments.
This movement planning segmentation has two major advantages.
First, it enables the network to consider feedback of the movement execution in the planning process and, second, the network can react to changing contexts, e.g., a changing target position.
Furthermore, it allows the network to update itself during planning, providing a mechanism for online model learning and adaptation to changing environments or constraints.
The general idea of how we enable the model to plan and adapt online is illustrated in Figure~\ref{fig:onlineFramework}.

To ensure a continuous execution of segments, the planning phase of the next segment needs to be finished before the execution of the current segment finished.
On the other hand, planning of the next segment should be started as late as possible to incorporate the most up-to-date feedback into the process.
Thus, for estimating the starting point for planning the next segment, we calculate a running average over the required planning time and use the three sigma confidence interval compared to the expected execution time.
The expected execution time is calculated from the distance the planned trajectory covers and a manually set velocity.
The learning part can be done right after a segment execution is finished.
The alignment of these processes are visualized in Figure~\ref{fig:onlineFramework}A.

As the recurrent network consists of stochastic spiking neurons, the network models a distribution over movement trajectories rather than a single solution. 
In order to create a smooth movement trajectory, we average over multiple samples drawn from the model when planning each segment.
Before the final mental movement trajectory is created by averaging over the drawn samples, we added a sample rejection mechanism.
As spiking networks can encode arbitrary complex functions, the model can encode multi-modal movement distributions.
Imagine that the model faces a known obstacle that can be avoided by going around either left or right.
Drawn movement samples can contain both solutions and when averaging over the samples, the robot would crash into the obstacle.
Thus, only samples that encode the same solution should be considered for averaging.

Clustering of samples could solve this problem, but as our framework has to run online, this approach is too expensive.
Therefore, we implemented a heuristic based approach that uses the angle between approximated movement directions as distance.
First a reference movement sample is chosen such that its average distance to the majority of the population is minimal, i.e., the sample that has the minimal mean distance to $90\%$ of the population is chosen as the reference. 
Subsequently only movement samples with an approximated movement direction close to the reference sample are considered for averaging.
As threshold for rejecting a sample, the three-sigma interval of the average distances of the reference sample to the closest $90\%$ of the population is chosen.

The feedback provided by the executed movement is incorporated before planning the next segment in two steps.
First, the actual position of the robot is used to initialize the sampling of the next segment such that planning starts from where the robot actually is, not where the previous mental plan indicates, i.e., the refractory state of the state neurons is set accordingly.
Second, the executed movement is used for updating the model based on the cognitive dissonance signal it generated.
In Figure~\ref{fig:onlineFramework}B this planning and adaptation process is sketched.

\subsection{Online Adaptation of the Recurrent Layer}
\label{sec:methodLearnig}
The online update of the spiking network model is based on the contrastive divergence (CD)~\cite{hinton2002training} based learning rules derived recently in~\cite{tanneberg16Deep}.
CD draws multiple samples from the current model and uses them to approximate the likelihood gradient.
The general CD update rule for learning parameters $\Theta$ of some function $f(x;\Theta)$ is given by
\begin{linenomath}
\begin{align}
\label{eq:CD_para_update}
\Delta\Theta &=  \left< \frac{\partial \log 
f(x;\Theta)}{\partial \Theta}\right>_{\mathbf{X}^{0}} \hspace{-7pt} - \left< \frac{\partial 
\log f(x;\Theta)}{\partial \Theta}\right>_{\mathbf{X}^{1}} \ ,
\end{align}
\end{linenomath}
where $\mathbf{X}^{0}$ and $\mathbf{X}^{1}$ denote the state of the Markov chain after $0$ and $1$ cycles respectively, i.e., the data and the model distribution.
We want to update the state transition function $\mathcal{T}(\mathbf{v}_t | \mathbf{v}_{t-1})$, which is encoded in the synaptic weights $\mathbf{w}$ between the state neurons (see Equation~\eqref{eq:transModel}).
Thus, learning or adapting the transition model means to change these synaptic connections.
The update rule for the synaptic connection $w_{k,i}$ between neuron $k$ and $i$ is therefore given by
\begin{linenomath}
\begin{align}
\label{eq:CDupdateW}
	w_{k,i} \leftarrow w_{k,i} + \alpha \Delta w_{k,i} \\ \text{with} \quad
	\Delta w_{k,i} = \tilde{v}_{t-1,k}\tilde{v}_{t,i} - \tilde{v}_{t-1,k}v_{t,i} \ \nonumber,
\end{align} 
\end{linenomath}
where $\tilde{v}$ denotes the spike encoding of the training data, $v$ the sampled spiking activity, $t$ the discrete timestep and $\alpha$ is the learning rate.
Here, we consider a resetting rectangular PSP kernel of one time step ($\tilde{v}_{t-1,k}$), a PSP kernel of $\tau$ time steps follows the same derivation and is used in the experiments.
In summary, this learning rules changes the model distribution slowly towards the presented training data distribution.
For a more detailed description of this spiking contrastive divergence learning rule, we refer to \cite{tanneberg16Deep}.
This learning scheme works for offline model learning when the previously gathered training data is replayed to an inhibitory initialized model.

For using the derived model learning rule in the online scenario, we need to make several changes.
In the original work, the model was initialized with inhibitory connections.
Thus, no movement can be sampled from the model for exploration until the learning process has converged.
This is not suitable in the online learning scenario, as a \textit{working} model for exploration is required, i.e., the model needs to be able to generate movements at any time.
Therefore, we initialize the synaptic weights between the state neurons using Gaussian distributions~\cite{stringer2002self}, i.e., a Gaussian is placed at the preferred position of each state neuron and the synaptic weights are drawn from these distributions with an additional additive negative offset term that enables inhibitory connections.
The synaptic weights are limited within $[-1, 1]$.

This process initializes the transition model with an uniform prior, where for each position, transitions in all directions are equally likely.
The variance of these basis functions and the offset term are chosen such that only close neighbors get excitatory connections, while distant neighbors get inhibitory connections, ensuring only small state changes within one timestep. i.e, a movement cannot \textit{jump} to the target immediately. 

Furthermore, the learning rule has to be adapted as we do not learn with an empty model from a given set of demonstrations but rather update a working model with online feedback.
Therefore, we treat the perceived feedback in form of the executed trajectory as a sample from the training data distribution and the mental trajectory as a sample from the model distribution in the supervised learning scheme presented in Equation~\eqref{eq:CD_para_update}.

\paragraph{Spike Encoding of Trajectories}
For encoding the mental and executed trajectories into spiketrains, Poisson processes with normalized Gaussian responsibilities of the state neurons at each timestep as time-varying input are used as in~\cite{tanneberg16Deep}. 
These responsibilities are calculated using the same Gaussian basis functions, centered at the state neurons preferred positions, as used for initializing the synaptic weights. 
More details on these responsibilities are given in Subsection~\ref{sec:localIMsignal} as they are also used for the local adaptation signals.
To transform these continuous responsibilities of the state neurons into binary spiketrains, they are scaled by a factor of $100$, limited into $[0, 10]$ and used as mean input to a Poisson distribution for each neuron.
The drawn samples for each neuron from these Poisson distributions for each timestep are compared to a threshold of $4$ and the neurons spike at time $t$ if this threshold is reached and the neuron has not spiked within its refractory period before.
The used parameters were chosen as they produced similar spiketrains as the ones sampled from the model.

\subsection{Global Intrinsically Motivated Adaptation Signal}
\label{sec:globalIM}
For online learning, the learning rate typically needs to be small to account for the noisy updates, inducing a long learning horizon, and thus requires a large amount of samples.
Especially, for learning with robots this is a crucial limitation as the number of experiments is limited.
Furthermore, the model should only be updated if \textit{necessary}.
Therefore, we introduce a time-varying learning rate $\alpha_t$ that controls the update step.
This dynamic rate can for example encode uncertainty to update only reliable regions, can be used to emphasize updates in certain workspace or joint space areas, or to encode intrinsic motivation signals.

In this work, we use an intrinsic motivation signal for $\alpha_t$ that is motivated by cognitive dissonance~\cite{festinger1962cognitive,kagan1972motives}.
Concretely, the dissonance between the mental movement trajectory generated by the stochastic network and the actual executed movement is used.
Thus, if the executed movement is similar to the generated mental movement, the update is small, while a stronger dissonance leads to a larger update.
In other words, learning is guided by the mismatch between expectation and observation.

This cognitive dissonance signal is implemented by the timestep-wise distance between the mental movement plan $\boldsymbol{\kappa}^{(m)}$ and the executed movement $\boldsymbol{\kappa}^{(e)}$.
Thus, the resulting learning factor is generated globally and is the same for all neurons.
As distance metric we chose the squared $L^2$ norm but other metrics could be used as well depending on, for example, the modeled spaces or environment specific features.
Thus, for updating the synaptic connection $w_{k,i}$ at time $t$, we change Equation~\eqref{eq:CDupdateW} to
\begin{linenomath}
\begin{align}
\label{eq:IMlearningRule}
	w_{k,i} \leftarrow w_{k,i} + \alpha_t \Delta w_{k,i} \\ 
	 \text{with} \quad \alpha_t = \lVert\boldsymbol{\kappa}_t^{(m)} - \boldsymbol{\kappa}_t^{(e)}\rVert_2^2 \nonumber \\
	\text{and} \quad \Delta w_{k,i} = \tilde{v}_{t-1,k}\tilde{v}_{t,i} - \tilde{v}_{t-1,k}v_{t,i} \quad \nonumber,
\end{align}
\end{linenomath}
where $\tilde{v}_t$ is the spike encoding generated from the actual executed movement trajectory $\boldsymbol{\kappa}^{(e)}_t$ and $v_t$ the encoding from the mental trajectory $\boldsymbol{\kappa}^{(m)}_t$ using the previously described Poisson process approach.

To stabilize the learning progress and for safety on the real system, we limit $\alpha_t$ in our experiments to $\alpha_t \in [0, 0.3]$ and use a learning threshold of $0.02$.
Thereby, the model update is only triggered when the cognitive dissonance is larger than this threshold, avoiding unnecessary computational resources, being more robust against noisy observations.
Note that during the experiments, $\alpha_t$ did not reach the safety limit and, therefore, the limit had no influence on the learning.
With this intrinsic motivated learning factor and the threshold that triggers adaptation, the update is regulated according to the model error and \textit{invalid} parts of the model are updated accordingly.
 
\subsection{Local Intrinsically Motivated Adaptation Signals}
\label{sec:localIMsignal}
In the previous subsection we discussed a mechanism for determining the cognitive dissonance signal that relies on the distance between the mental and the executed plan.
Thus, the resulting $\alpha_t$ is the same for all neurons at each timestep $t$, i.e., resulting in a \textit{global} adaptation signal.
Furthermore, the adaptation signal is calculated without taking the model into account.
To generate the adaptation signal incorporating the model, we need a different mechanism which is already inherent to the model.
Furthermore, we want to have individual learning signals for each neuron leading to a more focused and flexible adaptation mechanism.
Thus, the resulting learning signal should be \textit{local} and generated using the model.
To fulfill these properties, we utilize the mechanism that is already used in the model to encode trajectories into spiketrains -- the responsibilities of each neuron for a trajectory.
Inserting these individual learning signals into the update rule from Equation \eqref{eq:IMlearningRule} alters the update rule to
\begin{linenomath}
\begin{align}
\label{eq:localAdaptation}
	w_{k,i} \leftarrow w_{k,i} + \alpha_{t,i} \Delta w_{k,i} \\ 
	 \text{with} \quad \alpha_{t,i} = c ( \omega_{t,i}^{(m)} - \omega_{t,i}^{(e)})^2 \nonumber  \\
	\text{and} \quad \Delta w_{k,i} = \tilde{v}_{t-1,k}\tilde{v}_{t,i} - \tilde{v}_{t-1,k}v_{t,i} \quad \nonumber,
\end{align}
\end{linenomath}
with an additional constant scaling factor $c$.
For each neuron $i$, $\alpha_{t,i}$ encodes the time dependent  adaptation signal. 
These local adaptation signals are calculated as the squared difference between the responsibilities $\omega_{t,i}^{(m)}$ and $\omega_{t,i}^{(e)}$ for each neuron $i$ for the mental and the executed trajectory respectively.
These responsibilities emerge from the Gaussian basis functions centered at the state neurons positions that are also used for initializing the state transition model and the spike encoding of trajectories.
Therefore, the responsibilities are given by $\omega_{t,i}^{(m)} = b_i(\boldsymbol{\kappa}_t^{(m)})$ and $\omega_{t,i}^{(e)} = b_i(\boldsymbol{\kappa}_t^{(e)})$ with
\begin{linenomath}
\begin{align}
\label{eq:basisFunction}
	b_i(\bold{x}) = \exp \left( \frac{1}{2} (\bold{x} - \boldsymbol{p}_i)^\mathsf{T} \boldsymbol{\Sigma}^{-1} (\bold{x} - \boldsymbol{p}_i) \right) \quad , \nonumber 	
\end{align}
\end{linenomath}
where $\bold{p}_i$ is the preferred position of neuron $i$.
In the experiment we set $c = 3$, the learning threshold for $\alpha_{t,i}$ that triggers learning for each neuron to $0.05$ and limit the signal like in the global adaptation signal setting to $\alpha_{t,i} \in [0, 0.3]$.
Note, as in the global adaptation experiments, this limit was never reached in the local experiments and thus, had no influence on the results.

\subsection{Using Mental Replay Strategies to Intensify Experienced Situations}
As the encoding of trajectories into spiketrains using Poisson processes is a stochastic operation, we can obtain a whole population of encodings from a single trajectory.
Therefore, populations of training and model data pairs can be generated from one experience and used for learning.
We utilize this feature to implement a mental replay strategy that intensifies experienced situations to speed up adaptation.
In particular, we draw $20$ trajectory encoding samples per observation in the adaptation experiments, where each sample is a different spike encoding of the trajectory, i.e., a mental replay of the experienced situation.
Thus, by using such a mental replay approach, we can apply multiple updates from a single interaction with the environment.
The two mechanisms, using intrinsic motivation signals for guiding the updates and mental replay strategies to intensify experiences, lower the required number of experienced situations, which is a crucial requirement for learning with real robotic systems.

\figureGlobal 
\section{Results}
We conducted four experiments to evaluate the proposed framework for online planning and learning based on intrinsic motivation and mental replay.
In all experiments the framework had to follow a path given by way points that are activated successively one after each other.
Each way point remains active until it is reached by the robot.
In the first two experiments a realistic dynamic simulation of the KUKA LWR arm was used.
First, the proposed framework had to adapt to an unknown obstacle that blocks the direct path between two way points using the global adaptation signal and, second, by using the local adaptation signals and, third, by using constant learning rates (in combination with the global adaptation signal for triggering learning).
In the fourth experiment, we used a pre-trained model from the simulation in a real robot experiment to show that it is possible to transfer knowledge from simulation to the real system. 
Additionally, the model had to adapt online to a new unknown obstacle, again using the local adaptation signals, to highlight online learning on the real system.

\subsection{Experimental Setup}
\label{sec:expSetup}
For the simulation experiments, we used a realistic dynamic simulation of the KUKA LWR arm with a cartesian tracking controller to follow the reference trajectories generated by our model.
The tracking controller is equipped with a safety controller that stops the tracking when an obstacle is hit.
The task was to follow a given sequence of way points, where obstacles block the direct path between two way points in the adaptation experiments.
In the real robot experiment, the same tracking and safety controllers were used.
Figure~\ref{fig:darias} shows the simulated and real robot as well as the experimental setup.

By activating the way points successively one after each other as target positions using appropriate context neurons, the model generates online a trajectory tracking the given shape.
The model has no knowledge about the task or the constraint, i.e., the target way points, their activation pattern and the obstacle.
We considered a two-dimensional workspace that spans $[-1,1]$ for both dimensions -- the neuron's coordinate system -- encoding the $60 \times 60$ cm operational space of the robot.
Each dimension is encoded by $15$ state neurons, which results in $225$ state neurons using full population coding as in~\cite{tanneberg16Deep}.
The refractory period is set to $\tau = 10$, mimicking biological realistic spiking behavior and introducing additional noise in the sampling process.
The transition model is initialized by Gaussian basis functions centered at the preferred positions of the neurons (see \textit{Materials and Methods} for more details).
For the mental replay we used $20$ iterations, i.e., $20$ pairs of training data were generated for each executed movement.
All adaptation experiments were $300$ segments long, where $40$ trajectory samples were drawn for each segment and $10$ trials were conducted for each experimental setting.

\begin{table*}[thpb]
  \centering
  \begin{adjustbox}{max width=\textwidth}
   \begin{tabular}{c|c|c|c|c|c}
    & updates triggered ($\Downarrow$) & update time ($\Downarrow$) & planning time ($\Downarrow$) & exec. time ($\Uparrow$) & target reached ($\Uparrow$) \\
   \hline 
   global trigger & & & & & \\
   $\alpha = 0.001$ & $7.3 \pm 1.1$ & $42.6 \pm 4.2 ~ms$ & $ 0.238 \pm 0.047 ~s$ & $0.898 \pm 0.656 ~s$ & $10.5 \pm 0.5$ \\
   $\alpha = 0.01$ & $2.4 \pm 0.5$ & $43.7 \pm 4.2 ~ms$ & $ 0.237 \pm 0.046 ~s$ & $0.806 \pm 0.652 ~s$ & $8.5 \pm 3.2$ \\
   $\alpha = 0.1$ & $\mathbf{2.0 \pm 0.0}$ & $46.7 \pm 5.7 ~ms$ & $ 0.234 \pm 0.043 ~s$ & $0.771 \pm 0.670 ~s$ & $7.9 \pm 4.1$ \\
   \hline
   global $\alpha_t$ & $2.8 \pm 0.9$ & $43.3 \pm 4.1 ~ms$ & $ 0.241 \pm 0.044 ~s$ & $0.928 \pm 0.658 ~s$ & $10.4 \pm 0.8$ \\
   local $\alpha_{t,i}$ & $8.6 \pm 2.8$ & $52.6 \pm 7.5 ~ms$ & $ 0.235 \pm 0.042 ~s$ & $\mathbf{1.11 \pm 0.679 ~s}$ & $\mathbf{13.7 \pm 1.4}$ \\
   \end{tabular}
   \end{adjustbox}
   \caption{Evaluation of the adaptation experiments for $10$ trials with each the global, the local and constant learning signals in simulation.
   The values denote the number of times learning was triggered by a segment (updates triggered $=$ required samples $=$ physical interactions), the time required per triggered update including the mental replay strategy (update time), the planned execution time per segment (exec. time), the required time for planning a segment including sampling and post-processing (planning time), and the number of times the blocked target was reached within the budget of $300$ segments (target reached), i.e., number of times all way points were visited. All values denote mean and standard deviation. The $\Downarrow$ and $\Uparrow$ symbols denote if a lower or higher value is better respectively. Note that the constant $\alpha$ settings use the global adaptation signal $\alpha_t$ for triggering learning.}
       \label{table:comparison}
  \end{table*} 

\figureLocal
\subsection{Rapid Online Model Adaptation using Global and Local Signals}
\label{sec:expAdaptation}
In this experiment, we want to show the model's ability to adapt continuously during the execution of the planned trajectory.
A direct path between two successively activated way points is blocked by an unknown non-symmetric obstacle, which results in a discrepancy between the  planned and executed trajectory due to the interrupted movement.

\paragraph{Constant Learning Rates and the Importance of the Learning Threshold}
The main and starting motivation of the project was to enable online adaptation in the proposed stochastic recurrent network.
Therefore, we first created the framework for online planning (and adaptation -- see Figure~\ref{fig:onlineFramework}).
Afterwards we started experiments with online adaptation using the original learning rule (see Equation~\eqref{eq:CDupdateW}) and a constant learning rate $\alpha$.
We were not able to find a constant $\alpha$ for which the online learning was successful and stable, i.e., learning to avoid the obstacles and generating valid movements throughout the whole experiment.
With small learning rates, learning to avoid obstacles was successful, however, as the model is updated \textit{permanently} and in areas that are not affected by the environmental change, the model became unstable over time, resulting in a model, that was not able to produce valid movements anymore. 
The effect on the transition model using different constant learning rate is shown in Figure~\ref{fig:learningOverivewFixed}, which shows the \textit{unlearned} transition model that cannot produce valid movements (compare to Figures~\ref{fig:learningOverivewGlobal}~\&~\ref{fig:learningOverivewLocal}).

These insights gave rise to the idea of using adaptive learning signals in combination with a learning threshold to trigger learning only when an unexpected change is perceived.
With these mechanisms, successful and stable online adaptation of the stochastic recurrent network was possible.

Most closely related to our work are potential fields methods for motion planning and extensions to dynamic obstacle avoidance (see ~\cite{chiang2015path,erdem2012goal,lee2003artificial, barraquand1992numerical,ge2002dynamic} for example). 
All these approaches are deterministic models that consider obstacles through fixed heuristics of repelling potential fields.
In contrast, in our work we learn to avoid obstacles online through interaction by using the unexpected perceived feedback. 
In addition to the gradient based method in~\cite{chiang2015path}, we can learn to avoid obstacles with unknown shapes through the interactive online approach and static obstacles do not need to be known a priori. 
To evaluate the benefit of the dynamic online adaptation signals, we additionally compare to a baseline of our model using constant learning rates (with the adaptive global signal as learning trigger). 
This model can be seen as an extension of~\cite{chiang2015path} using stochastic neurons with the ability to adapt the potential field whenever an obstacle is hit.     

\paragraph{Online Adaptation Experimental Results}
The effect of the online learning process using intrinsically motivated signals is shown in Figure~\ref{fig:learningOverivewGlobal} and Figure~\ref{fig:learningOverivewLocal} for the global and the local signals respectively, where the mental movement trajectories, the adapted models and the adaptation signals $\alpha_t$ and $\alpha_{t,i}$ for three trials are shown.
Additionally we compare to using different constant learning rates $\alpha$, which use the global adaptation signal and its learning threshold to trigger learning (see the previous paragraph for why this is important), but using the constant $\alpha$ for the update.

With the proposed intrinsically motivated online learning, the model initially tries to enter the \textit{invalid} area but \textit{recognizes}, due to the perceived feedback of the interrupted movement encoded in the cognitive dissonance signals, the unexpected obstacle.
As a result the model adapts successfully and avoids the obstacle.
This adaptation happens efficiently from only $2.8 \pm 0.9$ physical interactions -- planned segments that hit the obstacle, which is equal to the number of samples required for learning -- with the global learning signal, where the planned execution time of one segment is $0.928 \pm 0.658$ seconds.
Moreover, the learning phase including the mental replay strategy takes only $43.3 \pm 4.1$ milliseconds per triggered segment.

Update and planning time with the local learning signals are similar, but adaptation is triggered $8.6 \pm 2.8$ times and the planned execution time is $1.11 \pm 0.679$ seconds. 
The increase of triggered updates is induced by the higher variability and noise in the individual learning signals, enabling more precise but also more \textit{costly} adaptation.
Still, the required samples -- triggered updates -- for successful adaptation reflect a sample efficient adaptation mechanism for a complex stochastic recurrent network.
The longer execution time indicates that the local learning signals generate more efficient solutions, as every segment covers a larger part of the trajectory, i.e., less segments are required resulting in a higher number for reaching the blocked target.
The local adaptation signals reached the blocked target $13.7 \pm 1.4$ times, which outperforms the other adaptation signals.
These results are summarized in Table~\ref{table:comparison}.
Thus, during the adaptation the global learning signals need fewer interactions, but the resulting solutions afterwards are less efficient.
The different effects of the global and local learning signals are discussed in more detail in Section~\ref{sec:expCompare}.

The results when using constant learning rates are summarized in Table~\ref{table:comparison} as well.
The best result was achieved with a learning rate of $\alpha = 0.001$, resulting in similar number of reached targets like the global adaptation signal (see also Figure~\ref{fig:compare}), but required almost as much updates -- i.e., samples -- as the local adaptation signals.
In addition to tuning this additional parameter, i.e., the constant learning rate, an adaptive signal for triggering learning is still required for successful and robust adaptation.
Moreover, when using the higher constant learning rates, the learning was unstable in some trials even with the adaptive trigger signals, i.e., after adaptation no valid movements could be sampled anymore. 

By adapting online to the perceived cognitive dissonances, the model generates new valid solutions avoiding the obstacle within seconds from few physical interactions (samples) with both learning signals.

\figureTransfer
\subsection{Transfer to and Learning on the Real Robot}
\label{sec:transerExp}
With this experiment we show that the models learned in simulation can be transferred directly onto the real system and, furthermore, that the efficient online adaptation can be done on a real robotic system.
Therefore, we adapted the simulated task of following the four given way points.
Additionally to the obstacles that were already present in simulation, we added a new unknown obstacle to the real environment. 
The setup is shown in Figure~\ref{fig:darias}B.
The framework parameters were the same as in the simulation experiment, except that the recurrent weights of the neural network were initialized with one trial of the simulation.
For updating the model the local learning signals were used and therefore the model was initialized with the 1st trial of the local signals simulation experiments (1st column in Figure~\ref{fig:learningOverivewLocal}A).
On average, an experimental trial on the real robot took about $5{:}30$ minutes (same as in simulation) and Figure~\ref{fig:learningOverviewRobot} shows the execution and adaptation over time.

As we started with the network trained in simulation, the robot successfully avoids the first obstacles right away and no adaptation is triggered before approach the new obstacle (Fig.~\ref{fig:learningOverviewRobot} first column).

After $15$ segments, the robot collides with the new obstacle and adapts to it within $7$ interactions (Figure~\ref{fig:learningOverviewRobot} second and third column).
The mismatch between the mental plan and the executed trajectory is above the learning threshold and the online adaptation is triggered and scaled with $\alpha_{t,i}$ (Figure~\ref{fig:learningOverviewRobot}B).

To highlight the efficient adaptation on the real system, we depicted the mental plan after $15$, $18$ and $300$ segments in Figure~\ref{fig:learningOverviewRobot}A.
For the corresponding segments, the cognitive dissonance signals show a significant mismatch that leads to the fast adaptation, illustrated in Figure~\ref{fig:learningOverviewRobot}B-C.
After the successful avoidance of the new obstacle, the robot performs the following task while avoiding both obstacles and no further updates are triggered.

\section{Discussion}
\label{sec:expCompare}
In this section we evaluate and compare the learning signals, the resulting models after the adaptation process, and the generated movements of the local and the global learning signals. 

\figureCompare
\subsection{Efficiency of the Learned Solutions}
Comparing the generated movements in Figure~\ref{fig:learningOverivewLocal}A to the movements generated with the global adaptation signal in Figure~\ref{fig:learningOverivewGlobal}A, the model using the local learning signals anticipates the learned obstacle earlier resulting in more natural evasive movements, i.e., more efficient solutions.
Here we define efficiency as the number of segments required to reach the blocked target.
As shown in Figure~\ref{fig:learningOverivewLocal}B-C, each neuron has a different learning signal $\alpha_{t,i}$ and therefore a different timing and scale for the adaptation, i.e., the neurons adapt independently in contrast to the global signal.
These individual updates enable a more flexible and finer adaptation, resulting in more efficient solutions.
As a result, when using the local adaptation signals, the model favors the more efficient solution on the right and chooses the left solution only in some trials at all after adaptation.
In contrast, this behavior never occurred in all ten trials with the global signal.
 
This efficiency can also be seen in Figure~\ref{fig:compare}, where the required segments to reach the blocked target are shown for the local signals, the global signal, a constant learning rate $\alpha = 0.001$, and without any adaptation.
Note that due to the stochasticity in the movement generation, the model can reach the block target without adaptation as well.
However, without adaption the obstacle is only avoided occasionally through the stochasticity in sampling the movements.

In Figure~\ref{fig:compare}A the mean and standard deviation of the required segments for reaching the blocked target are shown for $10$ trials with each setting over the complete $300$ segments in each trial.
All adaptation mechanisms outperform the model without adaptation, whereas the local signals perform better than the global signal and the constant learning rate.
Similar, in Figure~\ref{fig:compare}B the cumulated required segments for reaching the blocked target consecutively are shown for each trial together with the mean and standard deviation over the trials.
Note that, as all trials were limited to $300$ segments, the number of times the blocked target was reached differs in the different settings and trials (see Table~\ref{table:comparison}), depending on the efficiency of the generated movements, i.e., the amount of segments used.

\figureCompareHist
\subsection{Comparison of the Learning Signals}
To investigate the difference in the generated movements when using the global or local signals, we analyzed the corresponding learning signals $\alpha_t$ and $\alpha_{t,i}$.
This evaluation is shown in Figure~\ref{fig:compareHist}, where the magnitudes of the generated learning signals are plotted with their occurring frequency.
When looking at the right half of the histograms -- the $\alpha_t$ and $\alpha_{t,i}$ with lower magnitude --, both learning mechanisms produce similar distributions of the learning signals magnitude.
The main difference is the range of the generated signals, i.e., the local mechanism is able to generate stronger learning signals.
Even though the frequency of these bigger updates is low -- about $15 \%$ of the total updates --, they cover $30 \%$ of the total update mass, where update mass is calculated as the sum over all generated learning signals weighted by their frequencies. 
In contrast, the biggest $15 \%$ of the global learning signals cover $34 \%$ of the update mass and are all smaller than the biggest $15 \%$ of the local signals.

The ability to generate stronger learning signals in addition to the flexibility of individual signals, enables the local adaptation mechanism to learn models which generate more efficient solutions. 
The importance of the flexibility enabled by the individual learning signals is further discussed in the subsequent section.

\subsection{Spatial Adaptation}
Investigating the structure of the changes induced by the different learning signals, reveals a difference in the spatial adaptation and especially in the strength of the changes.
In the lower rows of Figure~\ref{fig:learningOverivewGlobal}A and Figure~\ref{fig:learningOverivewLocal}A the changes in the models are visualized with heatmaps showing the \textit{average change} of \textit{synaptic input} each state neuron receives, e.g., a value of $-0.03$ indicates that the corresponding neuron receives more inhibitory signals after adaptation.
Additionally, the \textit{average change} of \textit{synaptic output} of each state neuron is depicted by the scaled neuron sizes.

When the model adapts with the global signal (Figure~\ref{fig:learningOverivewGlobal}), the incoming synaptic weights of neurons with preferred positions around the blocked area are decreased -- the model only adapts in these areas.
The neurons around the constraint are inhibited after adaptation and, therefore, state transitions to these neurons get less likely.
This inhibition hinders the network to sample mental movements in affected areas, i.e., the model has learned to avoid these areas.
Due to the global signal, the learning is coarse and the affected area is spread larger than the actual obstacle.

In contrast, when adapting using the local signals (Figure~\ref{fig:learningOverivewLocal}), the structure of the changes in the model are more focused.
The strongest inhibition is still around the obstacle -- and stronger than with the global signal --, but much less changes can be found \textit{in front} of the obstacle. 
This concentration of the adaptation can also be seen when comparing the changes in the synaptic input and output.
Both learning mechanism produce a similar change in the output, but very different changes in the input, i.e., the neurons adapted with the local signals \textit{learned to focus} their output more precisely.

These stronger and more focused adaptations seem to enable the models updated with the local learning signals to generate more efficient solutions and favor the simpler path.

\subsection{Learning Multiple Solutions} 
Even though during the adaptation phase the model only experienced one successful strategy to avoid the obstacle, it is able to generate different solutions, i.e., bypassing the obstacle left or right, with both adaptation mechanisms.
Depending on the individual adaptation in each trial, however, the ratio between the generation of the different solutions differs.
Especially when using the local signals, the frequency of the more efficient solution is higher, reflecting the efficiency comparison in Figure~\ref{fig:compare}.

The feature of generating different solutions is enabled by the model's intrinsic stochasticity, the ability of spiking neural networks to encode arbitrary complex functions, the planning as inference approach and the task-independent adaptation of the state transition model.

\section{Conclusion} 
In this work, we introduced a novel framework for probabilistic online motion planning with an efficient online adaptation mechanism. 
This framework is based on a recent bio-inspired stochastic recurrent neural network that mimics the behavior of hippocampal place cells~\cite{rueckert16Recu,tanneberg16Deep}.
The online adaptation is modulated by intrinsic motivation signals inspired by \textit{cognitive dissonance} which encode the mismatch between mental expectation and observation.
Based on our prior work on the global intrinsic motivation signal~\cite{tanneberg2017online,tanneberg2017efficient}, we developed in this work a more flexible local intrinsic motivation signal for guiding the online adaptation.
Additionally we compared and discussed the properties of these two intrinsically motivated learning signals.
By combining these learning signals with a mental replay strategy to intensify experienced situations, sample-efficient online adaptation within seconds is achieved.
This rapid adaptation is highlighted in simulated and real robotic experiments, where the model adapts to an unknown environment within seconds from few interactions with unknown obstacles without a specified learning task or other human input.
Although requiring a few interactions more, the local learning signals learn more focused and are able to generate more efficient solutions -- less segments to reach the blocked target -- due to the high flexibility of individual learning signals. 

In contrast to~\cite{rueckert16Recu}, where the \textit{task-dependent context neuron input} was learned in a reinforcement learning setup, we update the state transition model, encoded in the recurrent state neurons connections, to adapt \textit{task-independently} with a supervised learning approach.
This sample-efficient and task-independent adaptation lowers the required expert knowledge and makes the approach promising for learning on robotic systems, for reusability and for adding online adaptation to (motion) planning methods.

Learning to avoid unknown obstacles by updating the state transition model encoded in the recurrent synaptic weights is a step towards the goal of recovering from failures.
One limitation to overcome before that, is the curse of dimensionality of the full population coding used by the uniformly distributed state neurons to scale the model to higher dimensional spaces.
In future work therefore, we want to combine this approach with the factorized population coding from~\cite{tanneberg16Deep} -- where the model's ability to scale to higher dimensional spaces and settings with different modalities was shown -- and learning the state neuron population~\cite{erdem2012goal}, in order to apply the framework to recover from failure tasks with broken joints~\cite{christensen2013distributed,cully2015robots}, investigating an intrinsic motivation signal mimicking the avoidance of arthritic pain~\cite{kulkarni2007arthritic,leeuw2007fear}. 

With the presented intrinsic motivation signals, the agent can adapt to novel environments by reacting to the perceived feedback.
For active exploration, and thereby \textit{forgetting} or finding novel solutions after failures, we plan to investigate intrinsic motivation signals mimicking curiosity~\cite{oudeyer2016intrinsic} in addition. 

As robots should not be limited in their development by the learning tasks specified by the human experts, equipping robots with such task-independent adaptation mechanisms is an important step towards autonomously developing and lifelong-learning systems.


\section*{Acknowledgments}
This project has received funding from the European Union's Horizon 2020 research and innovation programme under grant agreement No \#713010 (GOAL-Robots) and No \#640554 (SKILLS4ROBOTS).
 
\figureConst
\bibliography{nnbib}


\end{document}